%% file: main.tex
\documentclass[10pt,twocolumn,letterpaper]{article}

\usepackage{iccv}              

\usepackage{algorithm}
\usepackage{algpseudocode}
\usepackage{adjustbox}
\usepackage{multirow}
\usepackage{tcolorbox}
\usepackage{subcaption}


\newcommand{\drishtikon}{\textit{\textsc{DrishtiKon}}}

\definecolor{iccvblue}{rgb}{0.21,0.49,0.74}
\usepackage[pagebackref,breaklinks,colorlinks,allcolors=iccvblue]{hyperref}

\def\paperID{*****} 
\def\confName{ICCV}
\def\confYear{2025}

\title{
    \drishtikon{}: Visual Grounding at Multiple Granularities in Documents
}

\author{Badri Vishal Kasuba\\
\and
Parag Chaudhuri\\
\and
Ganesh Ramakrishnan\\
\and
\texttt{ \{badrivishalk, paragc, ganesh\}@cse.iitb.ac.in} \\
    Department of Computer Science and Engineering  \\
    IIT Bombay, India \\
}

\begin{document}
\maketitle

\begin{abstract}

    Visual grounding in text-rich document images is a critical yet underexplored challenge for Document Intelligence and Visual Question Answering (VQA) systems.
    We present \drishtikon{}, a multi-granular and multi-block visual grounding framework designed to enhance interpretability and trust in VQA for complex, multilingual documents.
    Our approach integrates multilingual OCR, large language models, and a novel region matching algorithm to localize answer spans at the block, line, word, and point levels.
    We introduce the Multi-Granular Visual Grounding (MGVG) benchmark, a curated test set of diverse circular notifications from various sectors, each manually annotated with fine-grained, human-verified labels across multiple granularities.
    Extensive experiments show that our method achieves state-of-the-art grounding accuracy, with line-level granularity providing the best balance between precision and recall.
    Ablation studies further highlight the benefits of multi-block and multi-line reasoning.
    Comparative evaluations reveal that leading vision-language models struggle with precise localization, underscoring the effectiveness of our structured, alignment-based approach.
    Our findings pave the way for more robust and interpretable document understanding systems in real-world, text-centric scenarios with multi-granular grounding support.
    Code and dataset are made available for future research.
\end{abstract}

\section{Introduction}
\label{sec:introduction}

Visual grounding refers to the task of accurately localizing regions in an image that correspond to a given natural language query or question~\cite{visual_grounding_survey}. In the domain of Document Intelligence and Visual Question Answering (VQA), visual grounding plays a pivotal role in linking textual answers to their supporting evidence within visually complex, text-rich document images—such as government circulars, official memos, and administrative forms. This spatial linkage not only facilitates interpretability and trust but also enables downstream applications that rely on the provenance of extracted information.

While visual grounding has shown significant progress in natural image understanding~\cite{visual_grounding_survey, vg_paper1, Rohrbach2015GroundingOT}, its application to document images remains underexplored. Document layouts are inherently distinct: they are dense, heterogeneous, often multilingual, and exhibit a variety of hierarchical structures.
These characteristics introduce unique challenges for grounding models, particularly in scenarios requiring high precision and accountability—such as compliance auditing, information verification, and explainable AI.
Manually tracing the source of an answer in such documents is labor-intensive and prone to human error, emphasizing the need for automated, fine-grained visual grounding methods tailored to Document VQA~\cite{docvqa, doccvqa}.
In contrast to natural images, where objects are typically grounded spatially, document images require grounding over structured textual units. In this context, the spatial provenance of an answer— \textbf{where} it appears on the document—is as critical as the answer content itself. Accurately identifying and localizing supporting regions enhances model transparency, facilitates human verification, and supports regulatory tasks such as information extraction, digital preservation, and legal compliance.

In this paper, we tackle the task of \textit{multi-granular visual grounding} in Document VQA. Given a document image and a question, our objective is to return both the textual answer and its precise supporting region(s) in the document at varying levels of granularity. To this end, we propose \textbf{\drishtikon{}}, a novel framework that synergizes multilingual OCR, large language models (LLMs), and a hybrid region matching strategy to enable accurate answer localization at multiple granularities—block, line, word, and point.
To evaluate our approach systematically, we construct a new benchmark, Multi-Granular Visual Grounding (MGVG) enabling robust evaluation of grounding methods and highlights the challenges posed by real-world administrative documents.
Our key contributions are as follows:
\begin{itemize}
    \item We introduce \textbf{\drishtikon}, a multi-granular visual grounding framework tailored for document VQA, supporting localization at block, line, word, and point levels.
    \item We design a robust hybrid region matching algorithm that integrates fuzzy text matching, length normalization, and semantic penalties to accurately localize multi-line and multi-block answers.
    \item We curate and release the \textbf{MGVG benchmark}, a new dataset of 70 government circulars with 509 diverse QnA pairs and human-annotated fine-grained spatial labels.
    \item We perform extensive experiments and ablation studies to quantify grounding performance across granularity levels, offering insights into precision-recall trade-offs in Document VQA with grounding.
    \item We benchmark leading vision-language models and highlight their limitations in spatial grounding, demonstrating the superiority of our alignment-based framework for structured document understanding.
\end{itemize}

\section{Related Work}
\label{sec:related_work}

While Multimodal Large Language Models (MLLMs) have advanced document understanding~\cite{docowl2,docllm,tilt,udop,layoutlmv3,layoutllm,docformerv2}, visual grounding in text-rich documents remains a crucial yet underexplored challenge for document intelligence. Recent works such as TGDoc~\cite{tgdoc}, DOGE-Bench~\cite{doge}, and TRIG-Bench~\cite{trig} have introduced benchmarks and models to address this gap. Table~\ref{tab:benchmark-comparison} compares these benchmarks in terms of methodologies, granularity, and synthetic data generation strategies.

TGDoc~\cite{tgdoc} enhances spatial awareness in MLLMs through instruction tuning on 99K PowerPoint slides and 12K GPT-4-generated dialogues, leveraging BLIP2~\cite{blip2}, PaddleOCR~\cite{paddleocr2020}, and GPT-4~\cite{gpt4} to align text recognition with spatial cues. DOGE-Bench~\cite{doge} introduces multi-granular grounding and referring tasks across charts, posters, and PDFs, using a synthetic pipeline (DOGE-Engine) to generate 700K training samples and a 4K evaluation benchmark, enabling fine-grained referencing at block, line, and word levels. In contrast, TRIG-Bench~\cite{trig} focuses on grounding in real-world, layout-heavy documents using a hybrid OCR-LLM-human pipeline, offering 90K synthetic and 800 manually curated samples to assess MLLM performance in spatially complex settings. Collectively, these benchmarks reveal the limitations of current MLLMs in handling dense layouts and highlight the need for improved spatial reasoning in document-centric tasks.

Despite these advances, several critical gaps remain. Existing benchmarks often have limited multilingual coverage, lack comprehensive human-verified annotations, and do not fully address multi-block and multi-line answer localization in complex, real-world documents. Moreover, most prior approaches rely heavily on synthetic data or focus on single-granularity tasks, which limits their applicability to the diverse document types encountered in practice.

Our work addresses these limitations by introducing a multi-granular, multilingual, and human-annotated benchmark, and by proposing a novel region matching algorithm that enables robust answer localization across multiple granularities. This positions our approach as a significant step forward in developing interpretable and trustworthy document understanding systems.

\begin{table}[h]
    \centering
    \begin{adjustbox}{width=\linewidth}
        \begin{tabular}{l c c c c}
            \toprule
            \textbf{Benchmark} & \textbf{Granularity} & \textbf{Multi-Block} & \textbf{OCR} & \textbf{LLM} \\
            \midrule
            TGDoc              & Single               & No                   & PaddleOCR    & GPT-4        \\
            DOGE-Bench         & Multi                & No                   & EasyOCR      & GPT-4o       \\
            TRIG-Bench         & Single               & Yes                  & PaddleOCR    & GPT-4o       \\
            \textbf{Ours}      & Multi                & Yes                  & —            & —            \\
            \bottomrule
        \end{tabular}
    \end{adjustbox}
    \caption{Comparison of text-rich visual grounding benchmarks.}
    \label{tab:benchmark-comparison}
\end{table}
\section{\drishtikon{} Methodology}
\label{sec:methodology}

\begin{figure*}[h]
    \centering
    \includegraphics[width=\textwidth]{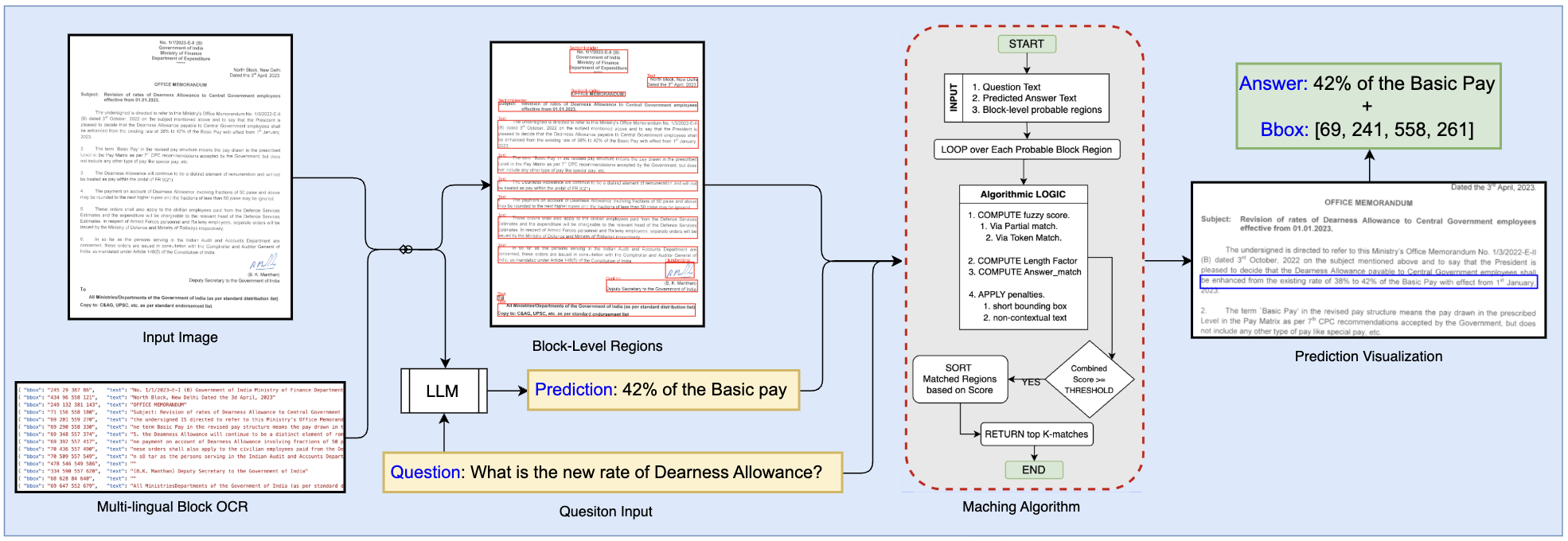}
    \caption{{\drishtikon}: Multi-granular Visual Grounding Pipeline with  matching algorithm flowchart on an example Illustration.}
    \label{fig:grounding_teaser}
\end{figure*}

This section details the proposed pipeline for \drishtikon{}, a multi-granular visual grounding solution for text-rich document images. Our approach is designed to address the unique challenges of document intelligence by enabling precise, interpretable answer localization across multiple levels of granularity. The framework integrates advanced multi-lingual OCR~\cite{doctr,suryaocr}, state-of-the-art large language models (LLMs)~\cite{llama3}, and a novel, hybrid scoring-based region matching algorithm. This combination allows for robust handling of complex layouts, multilingual content, and diverse document structures. Figure~\ref{fig:grounding_teaser} provides an overview of the \drishtikon{} pipeline.
Our methodology is designed for extensibility and can be adapted to a variety of document analysis tasks beyond VQA, including information extraction, compliance checking, and digital archiving. The following subsections describe each component of the pipeline in detail.

\subsection{Input Acquisition}
We begin with an input image of a text-rich document, typically a scanned official circular or memorandum, which forms the visual basis for question answering. The image is accompanied by a natural language question that queries for specific, extractable information, such as data, references, or contextual details present in the document.
The open-ended question allows for a range of possible answers spanning multiple text blocks within the document and is extractive, seeking specific phrases or data points in the image.

\subsection{Multi-Lingual Block-Level OCR}
The input image is processed using a robust multi-lingual OCR system using DocTR~\cite{doctr} and Surya-OCR~\cite{suryaocr} that segments the document into block-level text regions via Layout prediction.
Each region is encoded with its corresponding bounding box coordinates and transcribed content. This enables spatial decomposition of the document for fine-grained semantic grounding.

\subsection{Question-Conditioned Answer Prediction}
The extracted blocks, along with the user-input question, are fed into a large language model (LLM).
We use LlaMA-3.1-8B instruct open-source model~\cite{llama3} as our model, which performs semantic reasoning over the textual content and outputs a predicted answer.
This step simulates multimodal large language model (MLLM) inference in a purely textual modality, bypassing the need for visual input during answer generation by using OCR output as contextual information.

\subsection{Candidate Region Matching Algorithm}
\label{sec:grounding_region_matching}
To locate the predicted answer, we use a matching algorithm that iterates over each text block extracted by the OCR engine.
For each block, a composite matching score is calculated using the following components:

\begin{enumerate}
    \item \textbf{Fuzzy Score:} Computed using both the partial match and token-based matching similarity between the predicted answer and block content.
    \item \textbf{Length Factor:} Rewards blocks with reasonable length relative to the predicted answer with comparison of probable block matched text.
    \item \textbf{Penalty Functions:} Down-ranks blocks with either excessively short bounding boxes (indicative of noise) or lacking contextual semantic overlap.
\end{enumerate}

As described in the Algorithm~\ref{alg:region_matching}, Each candidate region is scored based on components which inturn have a scale of contribution, and those regions that scored above a pre-defined threshold are retained. The candidate blocks are sorted based on their composite scores. The top-k matched blocks are selected and visualized along with the predicted answer for interpretability.
This allows selection of the number of multi-block predictions based on the threshold parameters.

\begin{algorithm}
    \caption{Region matching algorithm for identifying probable block-level regions: Flowchart illustrating the steps involved in matching regions based on visual and textual features.}
    \label{alg:region_matching}
    \centering
    \begin{algorithmic}[1]
        \State \textbf{Input:}
        \State \hspace{1em}1. Question Text $Q$
        \State \hspace{1em}2. Predicted Answer Text $A$
        \State \hspace{1em}3. Block-level regions $R = \{r_1, r_2, \dots, r_n\}$

        \For{each region $r \in R$}
        \State \textbf{Step 1: Compute Fuzzy Score}
        \State \hspace{1em}a. Via partial match between $A$ and text in $r$
        \State \hspace{1em}b. Via token match between $A$ and text in $r$

        \State \textbf{Step 2: Compute Length Factor}
        \State \hspace{1em}Compute ratio or penalty based on bounding box/text length

        \State \textbf{Step 3: Compute Answer Match Score}
        \State \hspace{1em}Evaluate similarity between $A$ and content of $r$

        \State \textbf{Step 4: Apply Penalties}
        \State \hspace{1em}a. Penalize for short bounding boxes
        \State \hspace{1em}b. Penalize for non-contextual or irrelevant text

        \State \textbf{Step 5: Compute Combined Score}
        \If{Combined Score $\geq$ Threshold}
        \State Mark $r$ as valid match
        \EndIf
        \EndFor

        \State \textbf{Step 6: Sort} matched regions based on Combined Score
        \State \textbf{Step 7: Return} top $K$ matches

    \end{algorithmic}
\end{algorithm}

\subsection{Word-level Grounding Procedure}

\begin{figure}[h]
    \centering
    \begin{subfigure}{\linewidth}
        \centering
        \includegraphics[width=\linewidth]{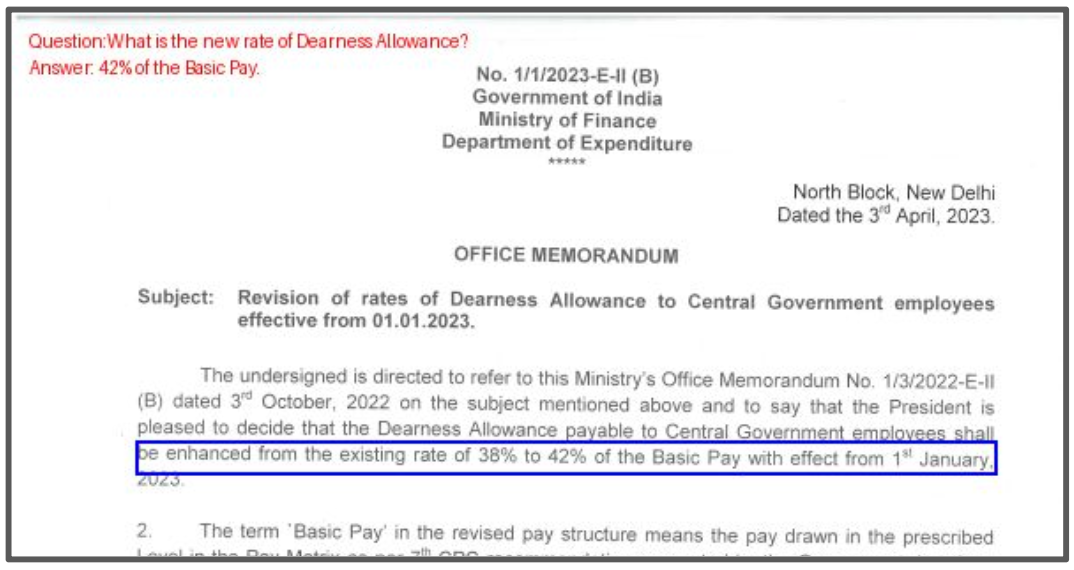}
        \caption{Step 1: Line-level grounding identifies the relevant line(s) containing the answer.}
        \label{fig:word_grounding_line}
    \end{subfigure}
    \vspace{0.5em}
    \begin{subfigure}{\linewidth}
        \centering
        \includegraphics[width=\linewidth]{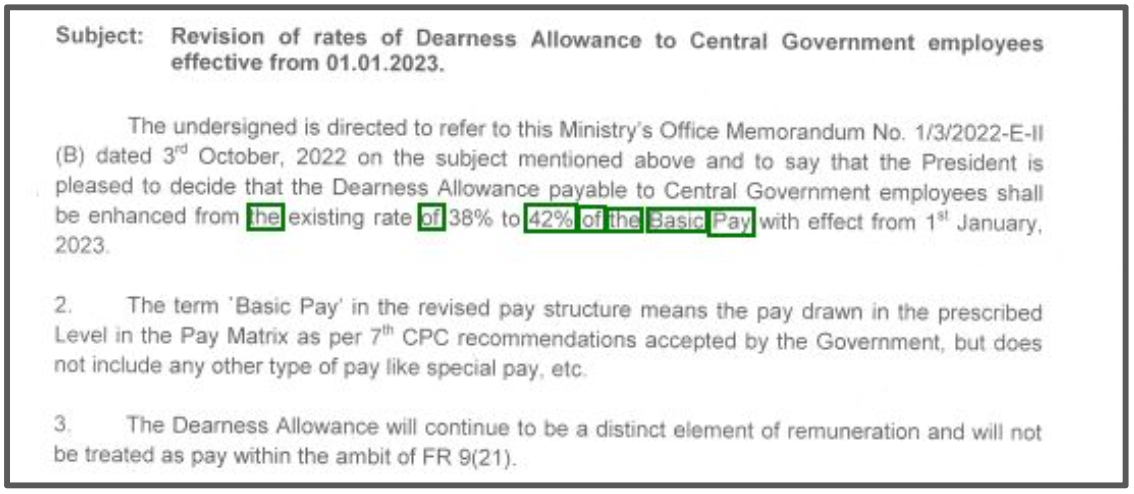}
        \caption{Step 2: Word-level matching highlights the individual words within the line(s) that correspond to the answer tokens.}
        \label{fig:word_grounding_match}
    \end{subfigure}
    \vspace{0.5em}
    \begin{subfigure}{\linewidth}
        \centering
        \includegraphics[width=\linewidth]{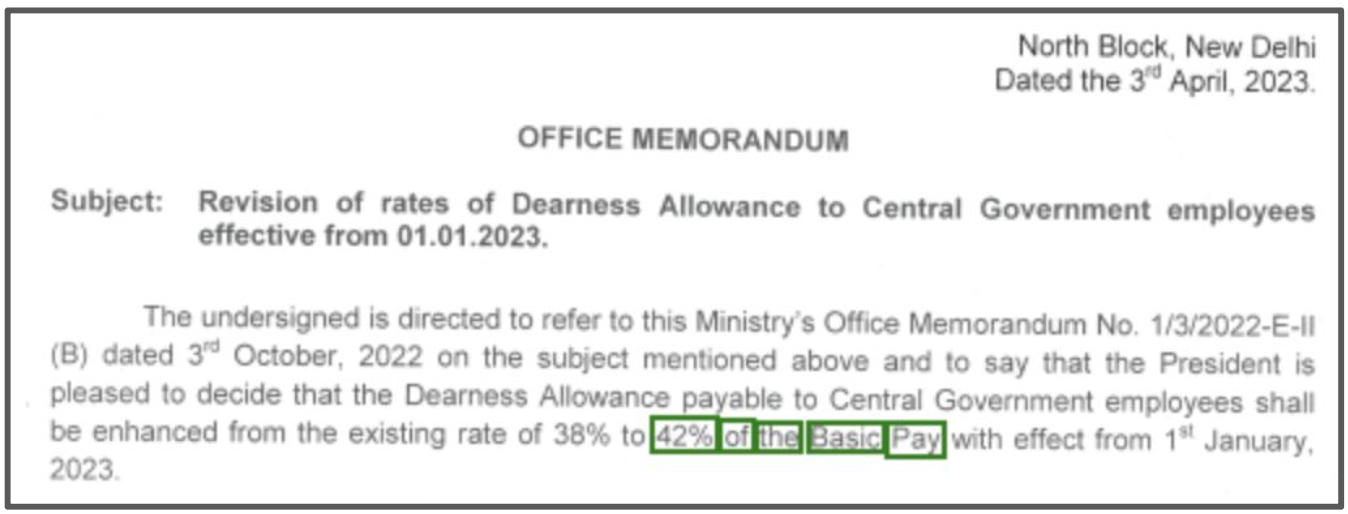}
        \caption{Step 3: Extraction of the longest contiguous sequence of matched words forms the final word-level grounded region.}
        \label{fig:word_grounding_final}
    \end{subfigure}
    \caption{Overview of the word grounding process.}
    \label{fig:word_grounding_overview}
\end{figure}

Word-level grounding is designed to achieve fine-grained localization of answer spans by identifying the precise words in a document image that correspond to the predicted answer. This process builds upon the results of line-level grounding and consists of the following steps:

\begin{enumerate}
    \item \textbf{Line-to-Word Mapping via Word Match:} Starting from the lines identified as relevant by the line-level grounding module (see Figure~\ref{fig:word_grounding_line}), each line is tokenized into individual words. For each predicted answer, a word-level matching operation is performed to locate all occurrences of the answer tokens within the candidate lines. This ensures that only those words directly contributing to the answer are considered for further localization (Figure~\ref{fig:word_grounding_match}).
    \item \textbf{Extraction of Longest Contiguous Word Indices:} To ensure semantic coherence and minimize fragmentation, the algorithm searches for the longest possible contiguous sequence of matched words that collectively form the answer span. This is achieved by identifying the maximal subsequence of word indices within the line(s) that matches the answer phrase, allowing for minor OCR or tokenization discrepancies. The resulting word-level bounding boxes are then aggregated to produce the final grounded region (Figure~\ref{fig:word_grounding_final}).
\end{enumerate}

This approach enables highly precise answer localization, supporting downstream applications such as information extraction, document redaction, and explainable VQA.

\section{Experiments and Results}

\subsection{Datasets}

To evaluate the effectiveness of our visual grounding framework, we curate a benchmark dataset derived from the Indic Government Circulars.
We sample 70 diverse document images encompassing complex layout structures and multi-lingual textual content, making them suitable for evaluating fine-grained spatial-semantic reasoning.
The dataset includes detailed annotations as summarized in Table~\ref{tab:dataset_stats}. As the TGDoc~\cite{tgdoc}, DOGE-Bench~\cite{doge} and TRIG-Bench~\cite{trig} datasets which are the existing Visual Grounding on text-centric images are yet not released and available as described in Section~\ref{sec:related_work} , we perform our experiments testing on the curated testset comprising of 509 QnA pairs with grounding.

\begin{table}[h]
    \centering
    \begin{tabular}{l c}
        \toprule
        \textbf{Attribute}          & \textbf{Count} \\
        \midrule
        Document Images             & 70             \\
        Question–Answer (QnA) Pairs & 509            \\
        Block-level regions         & 538            \\
        Line-level regions          & 988            \\
        Word-level regions          & 5,968          \\
        Point-level data            & 538            \\
        \bottomrule
    \end{tabular}
    \caption{Dataset statistics for Multi-Granular Visual Grounding (MGVG) test set: Distribution of instance counts across different granularity levels (block, line, word, point) within the annotated dataset.}
    \label{tab:dataset_stats}
\end{table}

\subsubsection{Annotation Tool}

\begin{figure}[h]
    \centering
    \includegraphics[width=0.8\linewidth]{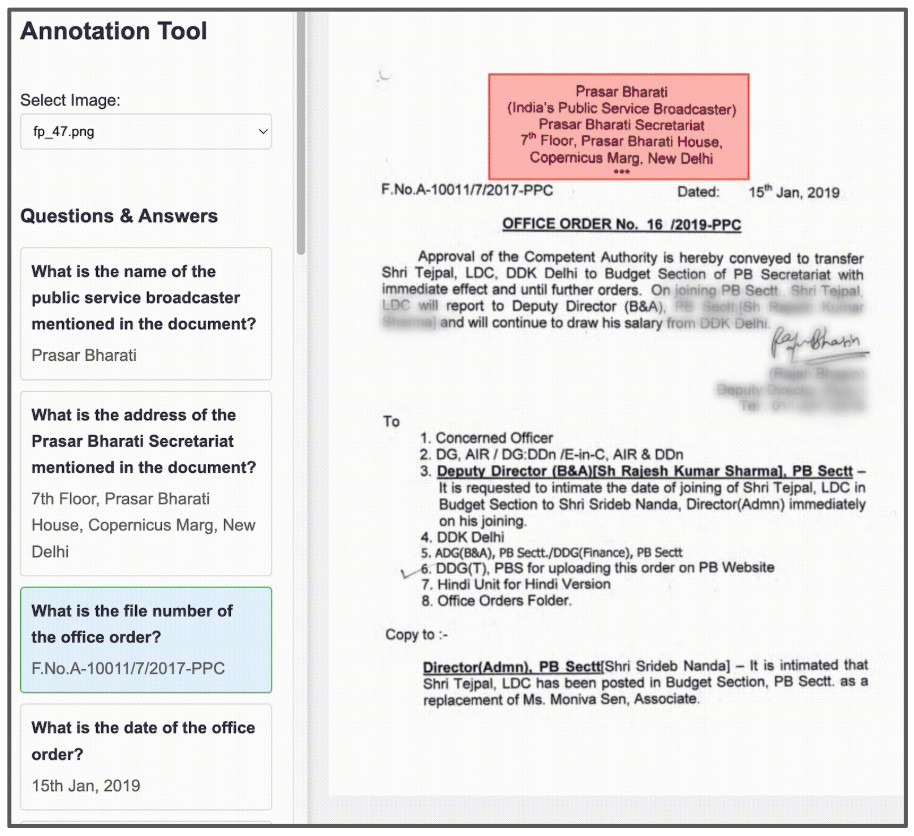}
    \caption{In-house annotation tool enabling multi-level region labeling: Tool interface showing options for defining and annotating regions of interest at varying levels of granularity.}
    \label{fig:annotation_tool}
\end{figure}

To construct the annotated test set used in our experiments, we developed a custom in-house annotation tool designed to support hierarchical and fine-grained annotation of document regions. This tool enables annotation at the block, line, word, and point levels, offering flexibility in defining answer spans that match natural visual layouts.
The tool has an interactive interface (Figure~\ref{fig:annotation_tool}) that allowed to label precise spans and relationships and also supports automatic text detection at word-level from the line-level annotated region aiding in enhanced annotation. Unlike other existing datasets~\cite{tgdoc,doge}, the dataset has been manually annotated instead of using synthetic data generation strategies via proprietary LLM.

\subsection{Granularity-wise Evaluation}

\begin{figure*}[h]
    \centering
    \includegraphics[width=\linewidth]{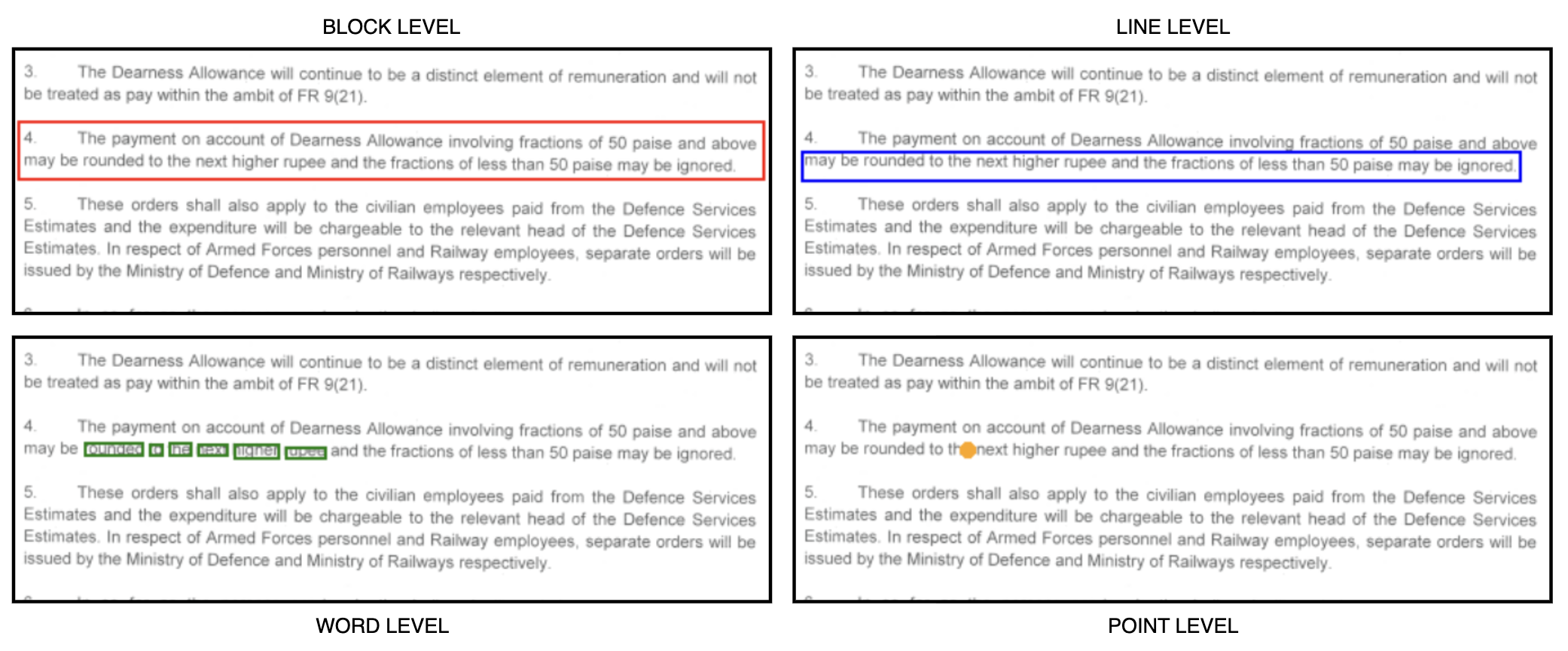}
    \caption{Visual grounding performance at varying granularities: Predicted bounding boxes show regions identified for question and answer pairs at block, line, word, and point levels.}
    \label{fig:grounding_granularity}
\end{figure*}

We evaluate the performance of our proposed visual grounding pipeline using standard metrics: \textbf{Precision}, \textbf{Recall}, and \textbf{F1-score}. These metrics are computed across four levels of grounding granularity: \textit{Block}, \textit{Line}, \textit{Word}, and \textit{Point}. The results, presented in Table~\ref{tab:evaluation_metrics}, correspond to the predictions generated using our Region Matching Algorithm under the \textit{Question + Ground Truth Answer} input.
Figure~\ref{fig:grounding_granularity} illustrates the visual grounding qualititative results across these granularities.

\begin{table}[h]
    \centering
    \begin{tabular}{lccc}
        \toprule
        \textbf{Level} & \textbf{Precision (\%)} & \textbf{Recall (\%)} & \textbf{F1-score (\%)} \\
        \midrule
        Block          & 53.89                   & 68.22                & 60.21                  \\
        Line           & \textbf{65.06}          & \textbf{73.68}       & \textbf{69.10}         \\
        Word           & 60.32                   & 49.87                & 54.60                  \\
        Point          & 60.66                   & 51.30                & 55.79                  \\
        \bottomrule
    \end{tabular}
    \caption{Grounding accuracy at different granularities}
    \label{tab:evaluation_metrics}
\end{table}

We observe that \textbf{Line-level grounding} achieves the highest F1-score of \textbf{69.10\%}, demonstrating that this granularity provides the best trade-off between precision and recall.
This is likely due to the structural nature of government circulars, where answers often span single, well-formed lines.
\textbf{Block-level grounding} benefits from capturing semantically complete units but suffers from region fragmentation and inconsistent bounding box aggregation.
Meanwhile, \textbf{word-level} and \textbf{point-level} grounding, although more fine-grained, are more susceptible to errors from OCR misalignment, which leads to reduced recall.
Overall, these findings suggest that mid-level granularities (especially lines) are optimal for visual grounding in structured documents.
However, as highlighted in recent work such as MOLMO~\cite{molmo}, point-level finetuning can still be beneficial for training as it provides required minimal grounding context and reducing training costs.





\subsection{Model-wise Evaluation}

We further compare the Region Matching Algorithm with multiple vision-language models (VLMs), including LLaMA 3.1~\cite{llama3} and Qwen2.5-VL~\cite{qwen2vl}.
All models are evaluated on the line-level grounding task under various supervision regimes: using both the question and predicted answer (\textit{Q+A}), using only the question, and hybrid configurations combining predicted answers with grounding algorithms.
The results are presented in Table~\ref{tab:line_level_grounding_ocr}.

\begin{table*}[h]
    \centering
    \resizebox{\linewidth}{!}{%
        \begin{tabular}{lccccccl}
            \toprule
            \textbf{Input (Text + BBox)} & \textbf{OCR (QA)} & \textbf{Model (QA)}  & \textbf{OCR (Grnd.)} & \textbf{Model (Grnd.)} & \textbf{P (\%)} & \textbf{R (\%)} & \textbf{F1 (\%)} \\
            \midrule
            Ground Truth Answer          & -                 & -                    & YES                  & Algorithm              & \textbf{65.06}  & \textbf{73.68}  & \textbf{69.10}   \\
            Ground Truth Answer          & -                 & -                    & YES                  & LLAMA                  & 49.20           & 64.98           & 56.00            \\
            Ground Truth Answer          & -                 & -                    & No                   & Qwen2.5VL              & 6.00            & 4.55            & 5.18             \\
            Predicted Answer             & YES               & LLaMA                & YES                  & Algorithm              & \textbf{43.97}  & \textbf{53.14}  & \textbf{48.12}   \\
            Predicted Answer             & YES               & LLaMA                & YES                  & LLAMA                  & 37.43           & 47.77           & 41.97            \\
            Predicted Answer             & NO                & PATRAM~\cite{patram} & YES                  & Algorithm              & 35.58           & 30.97           & 33.12            \\
            \bottomrule
        \end{tabular}%
    }
    \caption{Line-level grounding evaluation (10 max boxes, $IoU = 0.5$) with OCR-based input (text + bbox).}
    \label{tab:line_level_grounding_ocr}
\end{table*}

The Region Matching Algorithm, when paired with ground truth answers, outperforms all other methods and represents the skyline performance for line-level grounding tasks on this dataset under ideal conditions.

\textbf{Qwen2.5-VL}~\cite{qwen2vl}, despite being a general-purpose multimodal model, underperforms drastically across all metrics. This suggests its current limitations of OCR-free and Vision-Language Model (VLM) methods in handling text-centric visual grounding tasks.

Notably, with the provided OCR input context, \textbf{LLaMA 3.1} achieves competitive performance in predicting the visual grounding indicating that LLMs can effectively reason about spatial relationships when given structured textual context.

We also present a comparative evaluation at the block level in Table~\ref{tab:block_level_eval}. Once again, the Region Matching Algorithm maintains superior performance at the block level, emphasizing its consistent advantage in spatially grounded reasoning. The performance gap also reinforces the importance of layout-aware alignment in real-world document Question Answering (QA) settings.

\begin{table}[h]
    \centering
    \resizebox{\linewidth}{!}{%
        \begin{tabular}{lccc}
            \toprule
            \textbf{Method}             & \textbf{P (\%)} & \textbf{R (\%)} & \textbf{F1 (\%)} \\
            \midrule
            Algorithm                   & 53.89           & 68.22           & 60.21            \\
            LLaMA with Ground Truth     & 41.72           & 67.47           & 51.56            \\
            LLaMA with Predicted Answer & 32.88           & 54.46           & 41.01            \\
            \bottomrule
        \end{tabular}
    }
    \caption{Comparison of block-level grounding accuracy across different methods: Results are reported as IoU on the test set.}
    \label{tab:block_level_eval}
\end{table}

\subsection{Qualitative Comparison: Limitations of LLM-based and Algorithm-based Grounding}

\begin{figure*}[h]
    \centering
    \begin{subfigure}{0.43\linewidth}
        \centering
        \includegraphics[width=\linewidth]{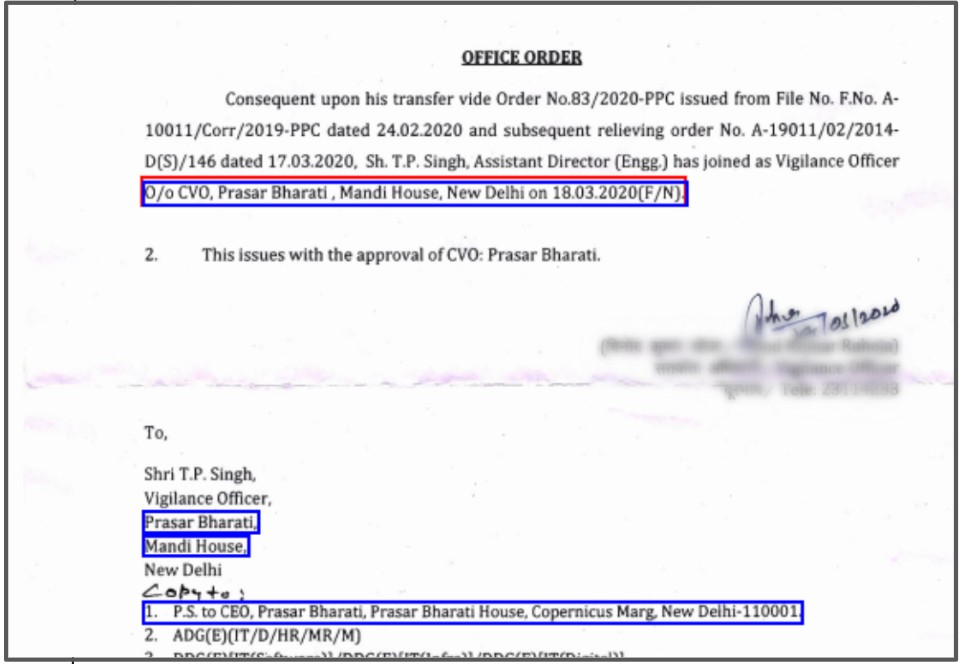}
        \caption{Algorithm-based grounding highlights multiple candidate regions but cannot reason about context, leading to over-selection.}
        \label{fig:algo_limitation}
    \end{subfigure}\hfill
    \begin{subfigure}{0.55\linewidth}
        \centering
        \includegraphics[width=\linewidth]{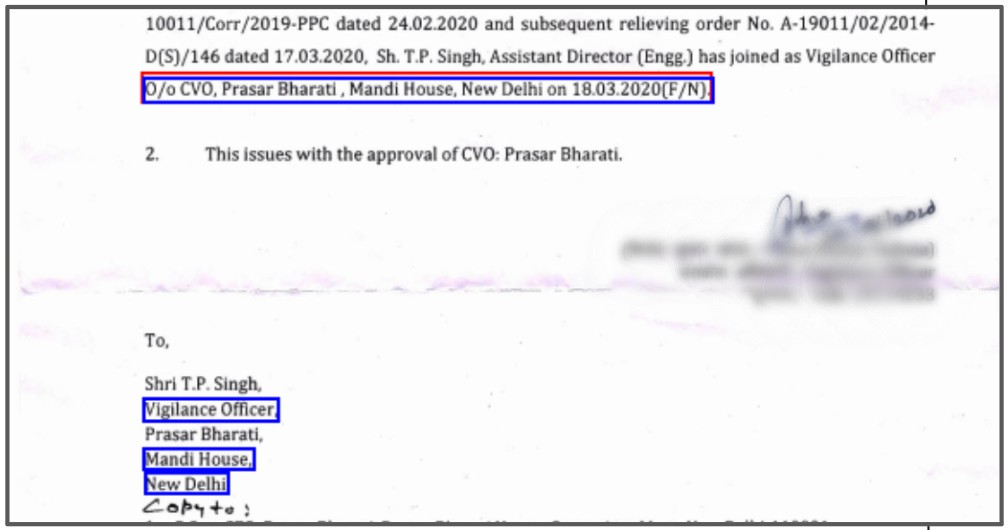}
        \caption{LLM-based~\cite{llama3} grounding predicts the answer but may fail to capture the full supporting content, especially for multi-region answers.}
        \label{fig:llm_limitation}
    \end{subfigure}
    \caption{Qualitative comparison of individual LLM-based and algorithm-based grounding (a) Algorithm-based method is precise in matching but lacks reasoning. (b) LLM-based approach can understand context but may miss spatial or structural details. This example demonstrates the complementary limitations of both approaches in Document VQA.}
    \label{fig:llm_algo_limitation}
\end{figure*}

Figure~\ref{fig:llm_algo_limitation} presents a qualitative comparison that highlights the complementary limitations of algorithm-based and LLM-based visual grounding approaches in document VQA. For the example question, "Where has Shri T.P. Singh been transferred to?", the algorithm-based method (Figure~\ref{fig:algo_limitation}) identifies multiple candidate regions by matching answer tokens and leveraging spatial cues. While effective in high-recall detection, it often results in over-selection, including redundant or irrelevant text regions due to its limited capacity for contextual reasoning. Conversely, the LLM-based method (Figure~\ref{fig:llm_limitation}), when provided with OCR-extracted text, demonstrates better semantic understanding of the question and can localize the correct answer span. However, it may fail to capture the complete supporting context, particularly when the answer spans multiple lines or regions, and is sensitive to OCR noise or incomplete textual input. This analysis underscores the inherent trade-offs between the two paradigms—algorithmic methods offer precise spatial grounding but lack semantic depth, while LLMs provide contextual understanding but often struggle with spatial alignment. These observations motivate the need for hybrid approaches that integrate structural precision with semantic reasoning for effective visual grounding in complex document understanding scenarios.

\subsection{Ablation Study}

To understand the impact of supporting multi-block and multi-line regions, we perform a detailed ablation study on the proposed region matching algorithm, varying the number of allowable blocks and lines during inference to enhance accuracy.

\textbf{Effect of Maximum Blocks}:\\
We evaluated performance while varying the maximum number of blocks from 1 to 5 in the region matching algorithm using question + ground truth answer alignment.
As shown in Table~\ref{tab:block_ablation}, allowing multiple blocks leads to improved recall at the expense of precision.
The optimal F1-score (62.68\%) is achieved when allowing up to 2 blocks, indicating that multi-block answers do exist but over-aggregation beyond two blocks introduces noise.

\begin{table}[h]
    \centering
    \begin{tabular}{cccc}
        \hline
        \textbf{Max Blocks} & \textbf{Precision} & \textbf{Recall} & \textbf{F1-score} \\
        \hline
        1                   & \textbf{71.83}     & 61.15           & 66.06             \\
        2                   & 58.67              & 67.29           & \textbf{62.68}    \\
        3                   & 55.57              & 67.66           & 61.02             \\
        4                   & 54.53              & \textbf{68.22}  & 60.61             \\
        5                   & 53.89              & \textbf{68.22}  & 60.21             \\
        \hline
    \end{tabular}
    \caption{Impact of the maximum number of blocks on region matching performance: Evaluation conducted on the test set, showing the relationship between the number of blocks and IoU.}
    \label{tab:block_ablation}
\end{table}

\textbf{Effect of Maximum Lines}:\\
We further extended the ablation over the number of lines allowed in the answer span.
Table~\ref{tab:line_ablation} and Figure~\ref{fig:line_trend} show the variation in precision, recall, and F1-score across iterations.
Precision drops steadily as more lines are aggregated, while recall consistently improves, reaching a plateau around 10 lines.
The F1-score stabilizes around 69.3\%, with the optimal performance achieved at 5 lines.

\begin{table}[h]
    \centering
    \begin{tabular}{cccc}
        \hline
        \textbf{Max Lines} & \textbf{Precision} & \textbf{Recall} & \textbf{F1-score} \\
        \hline
        1                  & \textbf{85.68}     & 41.80           & 56.19             \\
        2                  & 75.16              & 58.20           & 65.60             \\
        3                  & 70.52              & 66.09           & 68.23             \\
        4                  & 68.28              & 69.94           & 69.10             \\
        5                  & 66.89              & 71.96           & 69.33             \\
        6                  & 66.11              & 72.67           & 69.24             \\
        7                  & 65.79              & 73.38           & \textbf{69.38}    \\
        8                  & 65.44              & 73.58           & 69.27             \\
        9                  & 65.23              & \textbf{73.68}  & 69.20             \\
        10                 & 65.06              & \textbf{73.68}  & 69.10             \\
        \hline
    \end{tabular}
    \caption{Impact of the maximum number of lines on region matching performance: Results show the relationship between the maximum number of lines considered and the IoU score on a testset.}
    \label{tab:line_ablation}
\end{table}

\begin{figure}[h]
    \centering
    \includegraphics[width=\linewidth]{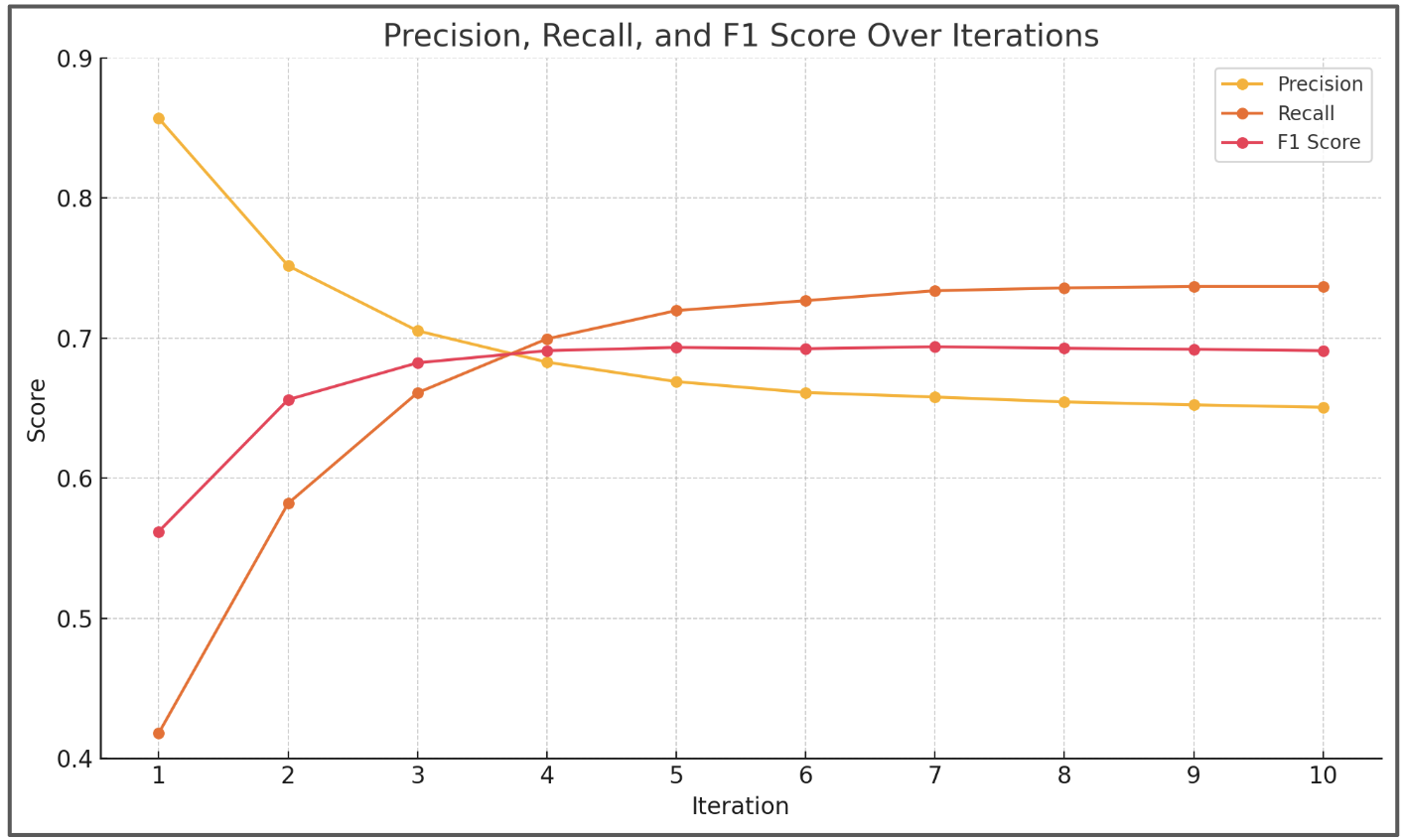}
    \caption{Trends in precision, recall, and F1-score with increasing maximum line thresholds at IoU = 0.5}
\end{figure}

This study highlights a crucial trade-off: precision dominates at lower aggregation levels, while recall improves as the answer spans grow larger. Allowing multi-line, multi-block reasoning significantly enhances performance, but must be carefully bounded to avoid degrading precision.
For our task, 2-block and 5-line aggregation offers the most balanced performance.

\subsection{Granularity and Multi-Block Trade-offs: Grounding Quality vs. Complexity}

The choice of grounding granularity and the handling of multi-block answers have a significant impact on both the quality of localization and the complexity of text similarity evaluation in document VQA.

\begin{figure}[h]
    \centering
    \includegraphics[width=\linewidth]{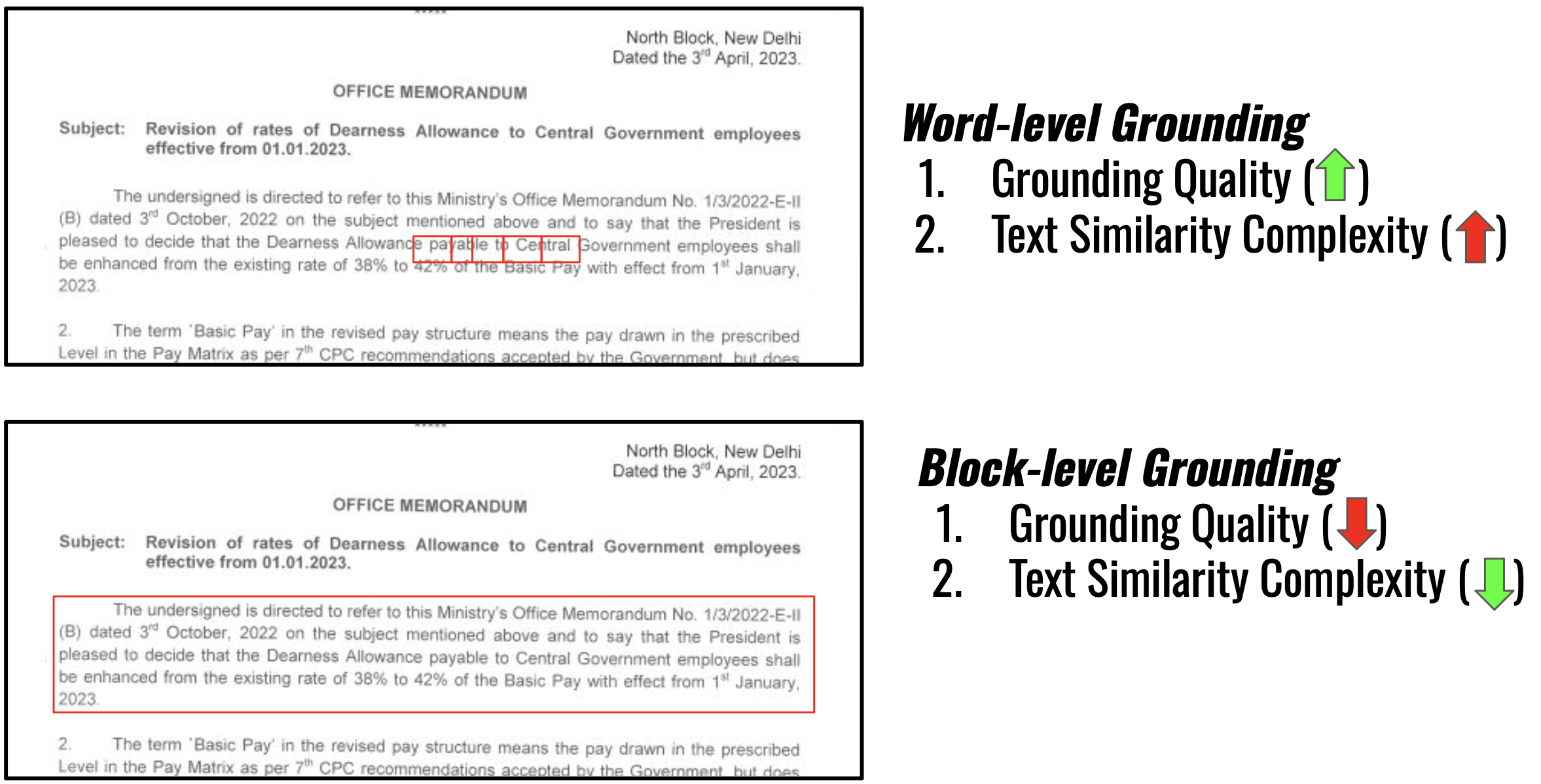}
    \caption{Comparison of grounding quality and text similarity complexity at different granularities. Word-level grounding achieves higher localization quality but increases text similarity complexity, while block-level grounding simplifies matching at the expense of precision.}
    \label{fig:granularity_quality_complexity}
\end{figure}

\paragraph{Granularity of Grounding:}
As illustrated in Figure~\ref{fig:granularity_quality_complexity}, word-level grounding enables highly precise localization of answer spans, resulting in superior grounding quality. However, this increased precision comes at the cost of higher complexity in text similarity computation, as the system must accurately match and align individual words, often in the presence of OCR noise or tokenization discrepancies. In contrast, block-level grounding simplifies the matching process by considering larger text regions, which reduces the complexity of text similarity evaluation but may lead to less precise localization and lower interpretability.

\begin{figure}[h]
    \centering
    \includegraphics[width=\linewidth]{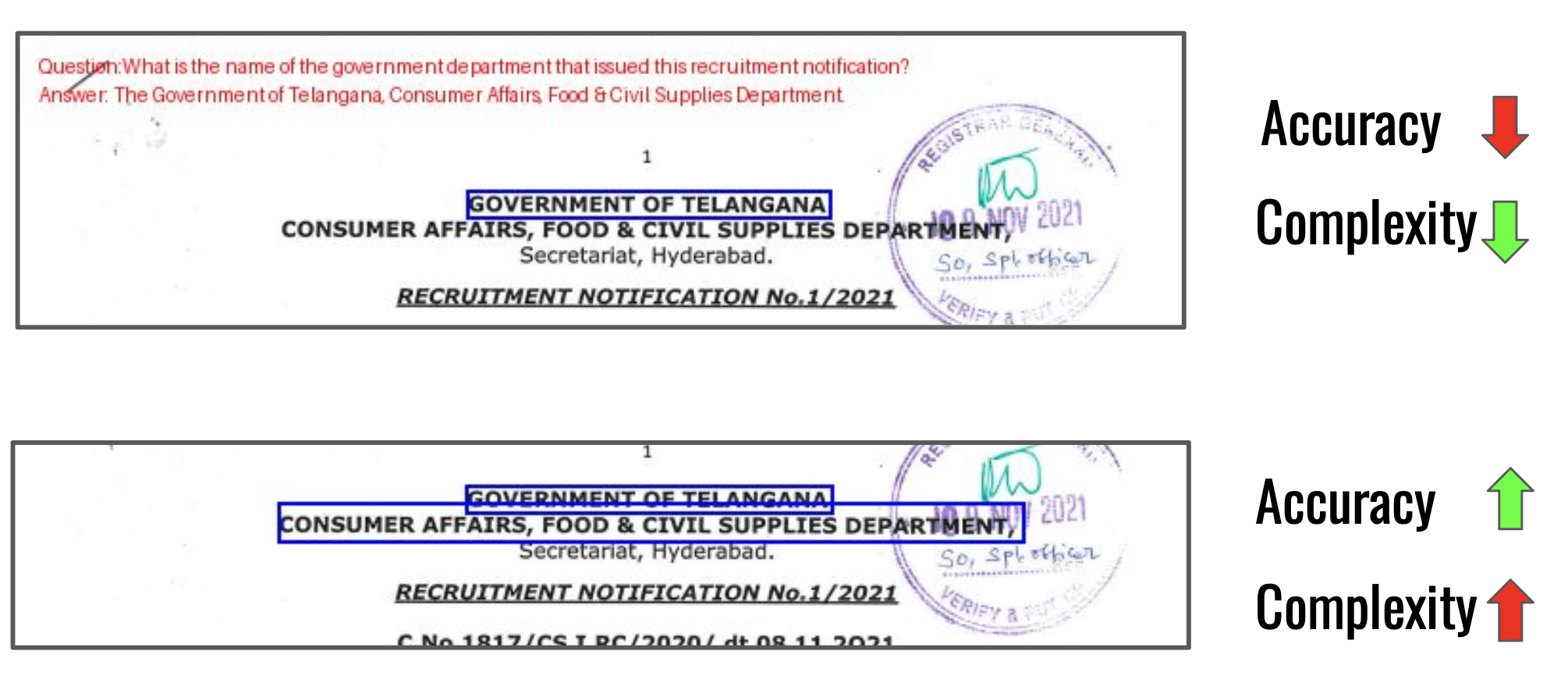}
    \caption{Impact of multi-block grounding on accuracy and complexity. Allowing multiple blocks can improve answer accuracy for complex queries, but increases the complexity of region selection and evaluation.}
    \label{fig:multi_block_grounding}
\end{figure}

\paragraph{Multi-Block Grounding:}
Document questions may require aggregating information from multiple, spatially separated blocks. As shown in Figure~\ref{fig:multi_block_grounding}, supporting multi-block grounding can improve answer accuracy for complex queries, but it also increases the complexity of both region selection and text similarity computation. When only a single block is considered, the process is simpler but may miss relevant information, reducing accuracy. Allowing multiple blocks to be selected increases the likelihood of capturing the complete answer, but introduces additional challenges in merging, ranking, and evaluating the candidate regions.

These trade-offs highlight the importance of selecting an appropriate grounding granularity and region aggregation strategy based on the specific requirements of the document VQA task. Our experiments quantitatively validate these observations, demonstrating that line-level grounding often provides the best balance between localization quality and computational complexity for structured documents.

\section{Conclusion}

In this work, we presented \textbf{\drishtikon}, a comprehensive multi-granular visual grounding framework aimed at enhancing the interpretability and reliability of VQA systems on text-rich document images. By integrating advanced multilingual OCR, large language models, and a novel region matching algorithm, our framework achieves accurate answer localization across multiple levels of granularity—block, line, word, and point.
To support rigorous evaluation, we introduced the Multi-Granular Visual Grounding (MGVG) benchmark. This benchmark provides a diverse set of human-annotated samples, enabling systematic assessment of visual grounding performance at varying granularities. Through extensive experiments and ablations, we show that mid-level granularities, especially line-level grounding, strike the optimal balance between precision and recall for document VQA. Our results further highlight the necessity of supporting multi-line and multi-block reasoning to handle the structural complexity inherent in real-world documents.
Our comparative analysis with state-of-the-art vision-language models reveals their limitations in fine-grained spatial localization, underscoring the advantages of structured, alignment-based approaches like \drishtikon. Together, our framework and benchmark establish a strong foundation for future research in transparent, accountable, and scalable document understanding systems.
Future work will focus on extending the framework to support multi-page and mixed-script grounding where the document is not a single page.
\paragraph{Broader Impact.}
The proposed framework holds significant potential for real-world applications such as automated compliance verification, digital recordkeeping, and reliable information extraction across government, legal, and enterprise settings. By improving transparency in automated document analysis, \drishtikon{} can contribute to more trustworthy AI systems and foster wider adoption of document intelligence technologies in critical decision-making workflows.
\paragraph{Privacy Considerations.}
To safeguard sensitive information and ensure ethical data handling, certain text regions in our document images, particularly those containing names or personally identifiable information, have been blurred. While many source documents originate from publicly available government circulars, we adopt a privacy-preserving approach to maintain responsible data sharing and protect individual and institutional confidentiality.

    {
        \small
        \bibliographystyle{ieeenat_fullname}
        \bibliography{main}
    }

\end{document}